# Mechanism Design for Cost Optimal PAC Learning in the Presence of Strategic Noisy Annotators

**Dinesh Garg**
IBM India Research Lab,
New Delhi, 110070 India
garg.dinesh@in.ibm.com

**Sourangshu Bhattacharya**
Yahoo! Labs,
Bangalore, 560071 India
sourangb@yahoo-inc.com

**S. Sundararajan**
Yahoo! Labs,
Bangalore, 560071 India
ssrajan@yahoo-inc.com

**Shirish Shevade**
Dept. of CSA, IISc
Bangalore, 560012 India
shirish@csa.iisc.ernet.in

## Abstract

We consider the problem of Probably Approximate Correct (PAC) learning of a binary classifier from noisy labeled examples acquired from multiple annotators (each characterized by a respective classification noise rate). First, we consider the complete information scenario, where the learner knows the noise rates of all the annotators. For this scenario, we derive sample complexity bound for the Minimum Disagreement Algorithm (MDA) on the number of labeled examples to be obtained from each annotator. Next, we consider the incomplete information scenario, where each annotator is strategic and holds the respective noise rate as a private information. For this scenario, we design a cost optimal procurement auction mechanism along the lines of Myerson's optimal auction design framework in a non-trivial manner. This mechanism satisfies incentive compatibility property, thereby facilitating the learner to elicit true noise rates of all the annotators.

## 1 Background

In supervised learning, it is usually assumed that the training set is sampled i.i.d. from some fixed distribution, and the true labels are readily available. In contrast to this, there are many real-world applications in which obtaining true labels is a time consuming or costly process. However, for such applications, acquiring non-expert labels is easy, fast, and inexpensive. Web based *crowdsourcing* [Howe, 2008] platforms like *Rent-A-Coder* and *Galaxy Zoo* allow any web-user to perform various data annotation tasks. Amazon's Mechanical Turk allow any individual to publish a crowdsourcing task. In such cases, labels obtained from such sources are typically noisy, as the annotators can be careless or even deceitful. Further, the annotators can act strategically if it helps them fetch better rewards.

The problem of learning with noisy labeled examples has been studied mostly in two different contexts: (1) studying the learnability of the problem and developing learning algorithms under the PAC learning framework [Valiant, 1984], [Angluin and Laird, 1988], [Aslam and Decatur, 1996], [Blum et al., 1994], [Decatur and Gennaro, 1995], [Decatur, 1997], [Kearns, 1993], [Littlestone, 1991], and (2) estimating the noise rates of the annotators and building robust classifier models (independent or joint estimation and learning) [Dawid and Skene, 1979], [Raykar et al., 2009], [Yan et al., 2010], [Donmez et al., 2010]. In the former context, most of the work has been done in a single noisy annotator scenario with different kinds of noise models including *Malicious Noise* [Valiant, 1985], [Kearns and Li, 1993], [Goldman and Sloan, 1995] and *Nasty Noise* [Bshouty et al., 2002]. In the latter context, the focus has been mainly using different classifier models and presenting different noise rate estimation techniques. Our work differs from these works in the following way: (1) we extend PAC learning sample complexity bound results to the multiple noisy annotators scenario as this problem is becoming more prevalent in recent times [Dekel and Shamir, 2009a], [Dekel and Shamir, 2009b], and (2) we design an optimal auction mechanism that facilitates eliciting (instead of estimating) true noise rates from the annotators followed by a cost-effective purchase of labeled examples satisfying the PAC learning constraint. We assume that annotators know their true noise rate. This is practical in many scenarios, e.g. many litigation service providing companies outsource document labeling tasks to paralegal agencies who in turn hire human editors. These agencies know the competence level of the editors through long standing relationship. The agencies bid for securing outsourcing contract while fully knowing the quality of its editors. The closest work in this direction is due to [Dekel et al., 2008] where they have focused on regression problem and took the elicitation approach via providing incentives to know the distribution information (privately held by the agents for evaluation). In contrast, our work is classification focused; furthermore, unlike the noisy annotators who charge labeling cost in our scenario, the strategic agents do not charge any price for annotation in their case. Instead, in their model, the agents have other vested interest in influencing the outcome of the learning; similar framework has also been used in [Meir et al., 2008], [Meir et al., 2009], [Dalvi et al., 2004], [L'Huillier et al., 2009], and [Kantarcioglu et al., 2008]. The main contributions of our work are:

- We consider the problem of PAC learning a binary classifier using noisy labeled examples obtained from multiple noisy annotators (in contrast to the conventional single noisy annotator scenario). We introduce the notion of annotation plan, and derive sample complexity bounds for PAC learning of finite concept class using the well-known minimum disagreement algorithm (MDA), in the known noise rates information scenario.

- We present an optimal auction mechanism to purchase labeled examples from strategic noisy annotators by eliciting true noise rate information in a more realistic scenario of unknown noise rates. Our approach is inspired by Myerson's Nobel prize winning work on optimal auction design [Myerson, 1981]. We derive allocation and payment rules for purchasing labeled examples at a near-optimal cost from strategic noisy annotators, satisfying the PAC constraint. To the best of our knowledge there has been no prior art with such results.

## 2 PAC Learning and Bounds

In this section, we provide basic definitions related to the PAC learning model [Valiant, 1984], [Angluin and Laird, 1988] with $n$ noisy annotators, and derive sample complexity bounds for PAC learning with $n$ noisy annotators. The PAC learning model comprises of an *instance space* $\mathscr{X}$ and a *concept class* $\mathscr{C}$. The instance space $\mathscr{X}$ is a fixed set which can be finite, countably infinite, $\{0,1\}^d$, or $\mathbb{R}^d$ for some $d \geq 1$. The concept class $\mathscr{C}$ is a set of *concepts*. A concept **c** is a subset of $\mathscr{X}$, which can equivalently be expressed as a boolean function from $\mathscr{X}$ to $\{0,1\}$, and it should be clear from the context whether **c** is referring to a subset or to a function. The task of the learner is to determine a close approximation to an unknown target (or true) concept $\mathbf{c}_t$, from the labeled examples. We assume that $\mathbf{c}_t \in \mathscr{C}$. The learner has access to $n$ noisy annotators as the sources of its training data. Each call to an annotator returns a labeled example $\langle x, y \rangle$, where instance $x \in \mathscr{X}$ is drawn randomly and independently according to some unknown (to the learner) sampling distribution $D$. The learner gets $m_i \geq 0$ labeled examples from annotator $i$, where $i = 1, \ldots, n$, which together constitute the training dataset. Finally, the learner employs a learning algorithm to output a hypothesis $h \in \mathscr{C}$, based on the training data. The annotator $i, (i = 1, \ldots, n)$ reports the label $y$ which is subject to an independent random mistake with a known probability $\eta_i$. So, the reported label is $y = \neg \mathbf{c}_t(x)$ with probability $\eta_i$ and $y = \mathbf{c}_t(x)$ with probability $(1 - \eta_i)$. This noise model is known as *random classification noise* and was first studied by [Angluin and Laird, 1988] and [Laird, 1988] for the single noisy annotator case. The probability $\eta_i, i = 1, \ldots, n$ is known as *noise rate* of the annotator $i$. In this paper, we assume that $0 < \eta_i < 1/2, i = 1, \ldots, n$.

## 2.1 PAC learning with $n$ noisy annotators

For any hypothesis $h \in \mathscr{C}$, the error rate (or generalization error) is defined to be the probability that $h(x) \neq \mathbf{c}_t(x)$ for an instance $x \in \mathscr{X}$ that is randomly drawn according to $D$. The error rate of a hypothesis $h$ is given by $\mathbb{P}^D(\mathbf{c}_t \Delta h)$, where $\mathbf{c}_t \Delta h \subseteq \mathscr{X}$ is the symmetric difference between sets $\mathbf{c}_t$ and $h$, and $\mathbb{P}^D(\cdot)$ is the probability of this event (calculated with respect to $D$). A hypothesis $h$ is said to be $\epsilon$-bad if its error rate is more than $\epsilon$, i.e. $\mathbb{P}^D(\mathbf{c}_t \Delta h) > \epsilon$. In the classical PAC model of [Valiant, 1984], the learner's goal is to come up with a learning algorithm which outputs an $\epsilon$-bad hypothesis $h$ with probability at most $\delta$, where the probability is defined with respect to the distribution of training examples of a fixed size. Such a learning algorithm is known as PAC learning algorithm. In general, the error rate of the hypothesis chosen by a learning algorithm critically depends on the number of training examples supplied to the algorithm. Thus, a learning algorithm with single annotator is said to satisfy PAC bound with respect to the *sample size* $m(\epsilon, \delta)$ if the following condition holds true: $\mathbb{P}^{m(\epsilon,\delta)}(\mathbb{P}^D(\mathbf{c}_t \Delta h) > \epsilon) < \delta$, where $h$ is the hypothesis output by the learning algorithm when trained on the $m(\epsilon, \delta)$ number of training examples. The sample size $m(\epsilon, \delta)$ is a non-negative integer valued function of the parameters $\epsilon$ and $\delta$. The probability $\mathbb{P}^{m(\epsilon,\delta)}(\cdot)$ is taken over the distribution of $m(\epsilon, \delta)$ training examples (noisy or non-noisy). For a given algorithm, the smallest sample size $m^*(\epsilon, \delta)$ for which it still satisfies PAC bound is known as its *sample complexity*. Now, we extend the PAC learning framework to the case of $n$ noisy annotators; starting with the following definitions:

- An **instance** of the PAC learning problem is a set of specifications of instance space $\mathscr{X}$, concept class $\mathscr{C}$, true concept $\mathbf{c}_t$, and sampling distribution $D$.

- An **annotation plan**, denoted by $\mathbf{m}(\epsilon, \delta) = (m_1(\epsilon, \delta), \ldots, m_n(\epsilon, \delta))$, is a vector of sample sizes (number of examples) annotated by the $n$ annotators. This quantity is analogous to sample size $m(\epsilon, \delta)$ in the single annotator case. In rest of the paper, we use $m_i$ and $m_i(\epsilon, \delta), i = 1, 2, \ldots, n$, interchangeably.

- A learning algorithm for $n$ noisy annotators is said to satisfy **PAC bound** for annotation plan $\mathbf{m} = (m_1, \ldots, m_n)$ if following holds true:

$$\mathbb{P}^{(m_1,\ldots,m_n)}(\mathbb{P}^D(\mathbf{c}_t \Delta h) > \epsilon) < \delta \quad (1)$$

Note that the noise rates $\eta_1, \ldots, \eta_n$ of the annotators could be very different. Hence the PAC bound depends not just on $\sum_{i=1}^n m_i$, but on the individual numbers $m_1, \ldots, m_n$ also. This motivates us to define the notions of **feasible and infeasible annotation plans**, as:

- For a given learning algorithm, an annotation plan $\mathbf{m} = (m_1, \ldots, m_n)$ is said to be *feasible* if the learning algorithm satisfies PAC bound (1) for every instance of the problem when training data is supplied as per this plan.

- Given an algorithm, an annotation plan $\mathbf{m} = (m_1, \ldots, m_n)$ is said to be *infeasible* if the algorithm **fails** to satisfy PAC bound (1) for at least one instance of the problem when training data is supplied as per this plan.

## 2.2 Feasible annotation plans for MDA

In this section, we consider a simple learning algorithm, namely *Minimum Disagreement Algorithm (MDA)* and derive PAC learnability bound on annotation plan complexity for this algorithm in the presence of $n$ noisy annotators. [Laird, 1988] analyzed this algorithm for single noisy annotator case. MDA outputs the hypothesis $h$, which minimizes the empirical loss, $L_e(h)$, on the training dataset. We describe MDA for multiple annotators, below.

**Algorithm 1 (MDA)** *Let $\mathscr{D} = \{\langle x_j^i, y_j^i \rangle \ i = 1, \ldots, n; \ j = 1, \ldots, m_i\}$ be the input training data, where $\langle x_j^i, y_j^i \rangle$ is supplied by annotator $i$ in $j^{th}$ call. The empirical loss $L_e(h)$ for hypothesis $h$ is given as:*

$$L_e(h) = \sum_{i=1}^n \sum_{j=1}^{m_i} \mathbf{1}(h(x_j^i) \neq y_j^i) \quad (2)$$

*where $\mathbf{1}(\cdot)$ is an indicator variable. Output hypothesis $h^* \in \mathscr{C}$, such that $L_e(h^*) \leq L_e(h), \forall h \in \mathscr{C}$ (use any tie breaking rule).*

Next we derive a characterization of feasible annotation plans for MDA. For this, we define a few events and their corresponding probabilities, assuming a finite concept class $\mathscr{C}$ having $|\mathscr{C}| = N < \infty$. The events are defined for an annotation plan $(m_1, \ldots, m_n)$ and a hypothesis $h$. We assume that $m_i$ samples of $x^i$ are drawn randomly and independently, according to the distribution $D$ by annotator $i$, and labels $y^i$ are flipped independently with noise rates $\eta_i$. The events $E_1, E_2, E_3$, and $E_4$, of our interest are defined as:

- $E_1(h, m_1, \ldots, m_n)$: The empirical error of a given hypothesis $h \in \mathscr{C}$ is no more than the empirical error of the true hypothesis $\mathbf{c}_t$, i.e. $L_e(h) \leq L_e(\mathbf{c}_t)$.

- $E_2(h, m_1, \ldots, m_n)$: The empirical error of a given hypothesis $h \in \mathscr{C}$ is the minimum across all hypotheses in the class $\mathscr{C}$, i.e. $L_e(h) \leq L_e(h') \; \forall h' \in \mathscr{C}$.

- $E_3(h, m_1, \ldots, m_n)$: MDA outputs a given hypothesis $h$.

- $E_4(\epsilon, m_1, \ldots, m_n)$: MDA outputs an $\epsilon$-bad hypothesis.

The probabilities of events $E_i, i = 1, 2, 3$ and $E_4$ are denoted by $\mathbb{P}^{(m_1, \ldots, m_n)}[E_i(h)]$ and $\mathbb{P}^{(m_1, \ldots, m_n)}[E_4(\epsilon)]$, respectively. Next, we show the following useful lemmas:

**Lemma 1** *Given a concept class $\mathscr{C}$ such that $\mathbf{c}_t \in \mathscr{C}$ and $N = |\mathscr{C}|$, the following holds true for any given annotation plan $(m_1, \ldots, m_n)$:*

$$\mathbb{P}^{(m_1, \ldots, m_n)}[E_4(\epsilon)] \leq (N-1) \times$$
$$\left[ \max_{h \in \mathscr{C}, h \text{ is } \epsilon\text{-bad}} \mathbb{P}^{(m_1, \ldots, m_n)}[E_1(h)] \right]$$

**Proof:** By definition of the events, for any hypothesis $h \in \mathscr{C}$ that is $\epsilon$-bad (for any $\epsilon > 0$), we have $E_3(h, m_1, \ldots, m_n) \subseteq E_2(h, m_1, \ldots, m_n) \subseteq E_1(h, m_1, \ldots, m_n)$. Also, $E_4(\epsilon, m_1, \ldots, m_n) = \bigcup_{h \in \mathscr{C}; h \text{ is } \epsilon\text{-bad}} E_3(h, m_1, \ldots, m_n)$. The lemma follows from taking probabilities of these events. Q.E.D.

Now, we state our main result regarding characterization of the feasible annotation plans for MDA.

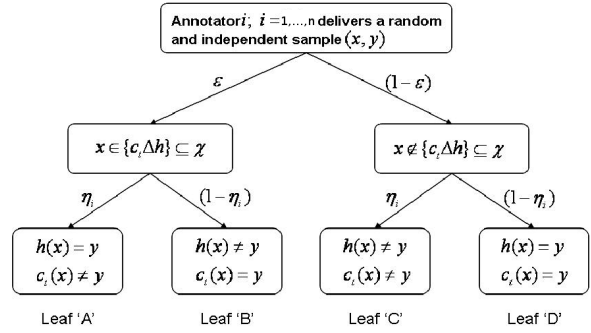

Figure 1: Probability Tree for $\mathbf{c}_t$ and $h$.

**Theorem 1** *Consider the PAC learning model with n noisy annotators and the MDA (Algorithm 1). Let $N = |\mathscr{C}| < \infty$. Then, for any given $0 < \epsilon, \delta < 1$ and $0 < \eta_i < 1/3, i = 1, \ldots, n$, if $m_i, i = 1, \ldots, n$, satisfy the following inequality then the MDA will satisfy PAC bound.*

$$\log(N/\delta) \leq \sum_{i=1}^{n} m_i \psi(\eta_i) \qquad (3)$$

*where $\psi(\eta_i) = \log\left[1 - \epsilon\left(1 - \exp\left(-(1-3\eta_i)/8\right)\right)\right]^{-1}$ for $i = 1, \ldots, n$*

*Remark:* The inequality (3) characterizes a subset of the feasible annotation plans. This characterization is independent of the problem instance and the tie breaking rule of the MDA.

**Proof:** MDA satisfies PAC bound iff there exists an annotation plan $(m_1, \ldots, m_n)$ such that $\mathbb{P}^{(m_1, \ldots, m_n)}[E_4(\epsilon)] < \delta$. From Lemma 1, it can be seen that for any $0 < \epsilon, \delta < 1$, if an annotation plan $(m_1, \ldots, m_n)$ satisfies the following condition, then MDA will satisfy PAC bound.

$$\left[ \max_{h \text{ is } \epsilon\text{-bad}} \mathbb{P}^{(m_1, \ldots, m_n)}[E_1(h)] \right] \leq \delta/N \qquad (4)$$

Notice a change in the LHS expression as compared to Lemma 1. The LHS in the above expression is an upper bound for the RHS of the expression in Lemma 1 (excluding $N-1$), because here $h$ may not belong to $\mathscr{C}$. This makes the bound independent of the problem instance, although MDA will only output $h \in \mathscr{C}$. Now, we upper bound the LHS of (4). To do this, we derive an upper bound for $\mathbb{P}^{(m_1, \ldots, m_n)}[E_1(h)]$ when hypothesis $h$ has an error rate of $\epsilon$ (for any $\epsilon \in (0, 1)$).

To derive this bound, we note that for any random and independent sample $(x, y)$ that is delivered by an annotator $i$, the probability of its agreeing (or disagreeing) with hypotheses $\mathbf{c}_t$ and $h$ (having error rate $\epsilon$) is given by a probability tree shown in Figure 1. From the tree, it can be seen that $\mathbb{P}^{(m_1,\ldots,m_n)}[E_1(h)]$ is same as the probability that the number of samples that fall under leaf B in the probability tree is at most the number of samples that fall under leaf A.

To compute the above quantity, first we compute the conditional probability, $\mathbb{P}^{k_1,\ldots,k_n}(L_e(h) \leq L_e(\mathbf{c}_t))$, defined as: if $k_i$ examples, $(0 \leq k_i \leq m_i)$, from annotator $i$ $(i = 1,\ldots,n)$ come from the set $(\mathbf{c}_t \Delta h)$, then the probability that empirical error of $h$ (given by $L_e(h)$) is less than or equal to empirical error of $\mathbf{c}_t$ (given by $L_e(\mathbf{c}_t)$).

Consider the random variable $Z_i^j, i = 1,\ldots,n$ $j = 1,\ldots,k_i$, which is the indicator of whether $j^{th}$ sample from $i^{th}$ annotator is from leaf node B, given that all the data points are from $\mathbf{c}_t \Delta h$ region. Thus, $\mathbb{P}(Z_i^j = 1) = (1 - \eta_i)$ and $\mathbb{P}(Z_i^j = 0) = \eta_i$. Let $Z = \sum_{i=1}^{n} \sum_{j=1}^{k_i} Z_i^j$. Then the event $L_e(h) \leq L_e(\mathbf{c}_t)$ is same as the event $Z \leq \sum_{i=1}^{n} k_i/2$. Hence, we are interested in finding an upper bound on $\mathbb{P}(Z \leq \sum_{i=1}^{n} k_i/2)$. We can use the multiplicative form of Chernoff bound (see e.g. Theorem 4.2 in [Motwani and Raghavan, 1995]), which says $\mathbb{P}[Z \leq (1 - \nu)\mu] \leq \exp(-\mu\nu^2/2)$, where $\mu = E[Z] = \sum_{i=1}^{n}(1 - \eta_i)k_i$. Hence, by letting $\nu = \frac{\sum_{i=1}^{n} k_i(1 - 2\eta_i)}{2\sum_{i=1}^{n} k_i(1 - \eta_i)}$, we get the following bound:

$$\mathbb{P}^{(k_1,\ldots,k_n)}(L_e(h) \leq L_e(\mathbf{c}_t)) \leq e^{\frac{-(\sum_{i=1}^{n} k_i(1 - 2\eta_i))^2}{8\sum_{i=1}^{n} k_i(1 - \eta_i)}}$$

Simplifying this bound, for $0 \leq \eta_i \leq 1/3$ we get:

$$\mathbb{P}^{(k_1,\ldots,k_n)}(L_e(h) \leq L_e(\mathbf{c}_t)) \leq e^{\frac{-\sum_{i=1}^{n} k_i(1 - 3\eta_i)}{8}} \quad (5)$$

Summing up the above conditional probability bound over all possible values of $k_i$, the total probability $\mathbb{P}^{(m_1,\ldots,m_n)}[E_1(h)]$ becomes:

$$\sum_{k_1=0}^{m_1} \cdots \sum_{k_n=0}^{m_n} \left( \prod_{i=1}^{n} \left( \binom{m_i}{k_i} \epsilon^{k_i}(1 - \epsilon)^{m_i - k_i} \right) \right. \quad (6)$$
$$\left. \mathbb{P}^{(k_1,\ldots,k_n)}(L_e(h) \leq L_e(\mathbf{c}_t)) \right)$$

Using the bound in (5), we get the following upper bound on $\mathbb{P}^{(m_1,\ldots,m_n)}[E_1(h)]$:

$$\prod_{i=1}^{n} \left( \sum_{k_i=0}^{m_i} \binom{m_i}{k_i} \epsilon^{k_i}(1 - \epsilon)^{m_i - k_i} \exp\left(-k_i(1 - 3\eta_i)/8\right) \right)$$

Using the moment generating function of the Binomial distribution, the bound becomes

$$\prod_{i=1}^{n} [1 - \epsilon(1 - \exp(-(1 - 3\eta_i)/8))]^{m_i}$$

In above bound, $h$ has an error rate exactly equal to $\epsilon$. However, this bound is valid for an $\epsilon$-bad hypothesis also because the expression decrease as $\epsilon$ increases. Substituting this upper bound on the LHS of (4), we get the desired claim. Q.E.D.

Note that the Theorem 1 is valid only for the range of $0 < \eta_i < 1/3$. However, we can extend the definition of $\psi(\cdot)$ to the boundary points in a manner that the same relation (3) holds true. For this, observe that minimum number of examples required from a single **non-noisy** annotator would be $m_0 = \log(N/\delta)/\log[1 - \epsilon]^{-1}$. This is because in such a case, we have $\mathbb{P}(L_e(h) \leq L_e(\mathbf{c}_t) \mid x \in \mathbf{c}_t \Delta h) = 0$ and $\mathbb{P}(x \notin \mathbf{c}_t \Delta h) = (1 - \epsilon)$. Hence, we can let $\psi(0) = \log[1 - \epsilon]^{-1}$. Also, we let $\psi(1/3) = \log[1 - \epsilon(1 - \exp(-(1/18)))]^{-1}$ and $m_{1/3} = \log(N/\delta)/\psi(1/3)$ from [Laird, 1988].

## 3 Cost Optimal Mechanism Design for PAC Learning

We consider the problem of procuring a feasible annotation plan when the learner needs to pay annotators for their efforts, under known and unknown noise rate scenarios. In the unknown noise rate scenario, we propose an auction model and present an optimal auction mechanism.

We assume that each annotator $i$ (with noise rate $\eta_i$) incurs an internal cost $c(\eta_i)$ of annotation for labeling one data point; note that the cost is dependent on the noise rate, and the cost function is same for all the annotators. The cost function is assumed to be a bounded, continuously differentiable, strictly decreasing function in $0 \leq \eta_i < 1/2 \; \forall \; i = 1,\ldots,n$. If an annotator is more competent (i.e. less noisy) then he can make more money by selling his services and time to somewhere else in the market, which translates to saying that his internal cost of annotation is high.

Consider a simplistic scenario of complete information where the learner knows noise rates of all the annotators. In such a case, the goal of the learner is purchase an annotation plan $m =$

$(m_1, m_2, \ldots, m_n)$ in such a way that the procurement cost (that is, cost of annotation), given by $\sum_{i=1}^{n} m_i c(\eta_i)$, is minimized subject to the PAC learning constraint (3). This is an integer linear programming problem. An approximate solution can be obtained by relaxing the integer constraint and rounding off the optimal solution to the nearest integer value.

Now, let us consider a more realistic scenario of incomplete information where the learner does not know noise rates $\eta = (\eta_1, \ldots, \eta_n)$. There are two possible approaches: (1) estimation, and (2) elicitation. In the estimation approach, the learner estimates $\eta_i$ using previously acquired examples (say, for example, comparing labels from different annotators). In the elicitation approach, the learner gets $\eta_i$ directly from the annotators. The former approach has the disadvantage that poor estimates result in either paying more (when overestimated) or not satisfying the PAC bound (when underestimated). Due to this reason, we are interested in elicitation. In this approach, the learner pays an incentive (a.k.a. price of information) to get $\eta_i$ from the annotators. Note that the learner needs to pay this price of information to elicit true noise rates. (Otherwise, the annotators can falsely report the noise rate.) For this purpose, we propose to design a procurement auction mechanism to procure a feasible annotation plan with minimum cost; now, the procurement cost also includes the price of information. This problem is challenging because from annotator's perspective, he would like to maximize his utility (i.e., the payment received minus the internal cost for annotation). The choice of mechanism depends crucially on various design parameters such as $N, \epsilon, \delta, \eta_i, c(\cdot)$, and the choice of the learning algorithm. We assume that $N, \epsilon, \delta, c(\cdot)$, and the choice of the learning algorithm are public knowledge, and only $\eta_i$ is the private information of $i^{th}$ annotator.

### 3.1 Procurement Auction Model

The learner solicits simultaneous and confidential bids for the noise rates from annotators. Let $\hat{\eta}_i$ be the bid of $i^{th}$ annotator that can possibly be a false noise rate. Assume that annotator $i$ draws his true noise rate $\eta_i$ in an independent random manner using a density function $\phi_i$ in the interval $I_i = [0, 1/3]$ with the corresponding cumulative distribution function $(\Phi_i)$, and let $\phi_i(\eta_i) > 0$ for all $\eta_i \in I_i$ and $i = 1, 2, \ldots, n$. Let $I = I_1 \times I_2 \times \ldots \times I_n$ and $\phi = \phi_1 \times \phi_2 \ldots \times \phi_n$ denote respective joint spaces. We use the subscript $-i$ to exclude $i^{th}$ annotator in any variable (e.g. $I_{-i}, \eta_{-i}$) and, we also use the notation $\hat{\eta} = (\hat{\eta}_i, \hat{\eta}_{-i})$.

After receiving the bids (i.e. $\hat{\eta} = (\hat{\eta}_1, \ldots, \hat{\eta}_n)$), the learner allocates a contract of supplying certain number of labeled examples to each annotator and an associated payment. Thus, a procurement auction mechanism is a pair of mappings $\mathcal{M} = (a, p)$, where $a : I \mapsto \mathbb{N}_0^n$ [1] is the **allocation rule** and $p : I \mapsto \mathbb{R}^n$ is the **payment rule**.

Given an auction mechanism $\mathcal{M} = (a, p)$, an annotator $i$, having noise rate $\eta_i$, gets the following utility when all the annotators report their bids $\hat{\eta}$:

$$u_i(\hat{\eta}; \eta_i) = p_i(\hat{\eta}) - a_i(\hat{\eta}) c(\eta_i) \qquad (7)$$

Note that the first and the second term denote the payment received from the learner and the internal cost in supplying the labeled examples, respectively. Since each annotator $i$ does not know $\eta_{-i}$ and moreover, others' bids $\hat{\eta}_{-i}$ affect his utility, it is useful to define **expected allocation rule** $\alpha$ and the **expected payment rule** $\pi$ for any mechanism $\mathcal{M} = (a, p)$ in the following manner (from $i^{th}$ annotator's perspective).

$$\alpha_i(\hat{\eta}_i) = \int_{I_{-i}} a_i(\hat{\eta}_i, \hat{\eta}_{-i}) \phi_{-i}(\hat{\eta}_{-i}) d\hat{\eta}_{-i} \qquad (8)$$

$$\pi_i(\hat{\eta}_i) = \int_{I_{-i}} p_i(\hat{\eta}_i, \hat{\eta}_{-i}) \phi_{-i}(\hat{\eta}_{-i}) d\hat{\eta}_{-i} \qquad (9)$$

The **expected utility** of annotator $i$, when he bids $\hat{\eta}_i$ while having true value $\eta_i$, can now be given by

$$U_i(\hat{\eta}_i; \eta_i) = \pi_i(\hat{\eta}_i) - \alpha_i(\hat{\eta}_i) c(\eta_i) \qquad (10)$$

When both arguments in (10) are same, we use $U_i(\eta_i)$ to mean $U_i(\eta_i; \eta_i)$ (for notational simplicity). Given this background, we first present several definitions that are essential to prove our results. A Mechanism $\mathcal{M} = (a, p)$ is said to be:

---

[1] The symbol $\mathbb{N}_0$ denotes the set of natural numbers inclusive of zero. The allocation and the payment rules are functions of $N, \epsilon, \delta, c(\cdot)$, and the algorithm. For notational simplicity, we drop these parameters.

- **Dominant Strategy Incentive Compatible** (DSIC) if for every annotator $i$ and for every possible true noise rate $\eta_i \in I_i$, the utility $u(\cdot)$ is maximized when $\hat{\eta}_i = \eta_i$ irrespective of what others are bidding, i.e., $u_i(\eta_i, \hat{\eta}_{-i}; \eta_i) \geq u_i(\hat{\eta}_i, \hat{\eta}_{-i}; \eta_i) \ \forall \ \hat{\eta}_i \in I_i, \hat{\eta}_{-i} \in I_{-i}$.
- **Bayesian Incentive Compatible** (BIC) if for every annotator $i$ and for every possible true noise rate $\eta_i \in I_i$, the expected utility $U_i(\cdot)$ is maximized when $\hat{\eta}_i = \eta_i$, i.e., $U_i(\eta_i) \geq U_i(\hat{\eta}_i; \eta_i) \ \forall \hat{\eta}_i \in I_i$. Note, any mechanism satisfying DSIC will also satisfy BIC but the other way is not necessarily true.
- **PAC compatible** if the annotation plan procured by this mechanism satisfies the PAC bound condition (3) whenever all the annotators report their true noise rates, i.e., $\log(N/\delta) \leq \sum_{i=1}^n a_i(\eta)\psi(\eta_i)$.
- **Individually Rational** (IR) if no annotator loses (in expected sense) anything by reporting true noise rates, i.e., $\pi_i(\eta_i) - \alpha_i(\eta_i)c(\eta_i) \geq 0 \ \forall \ \eta_i \in I_i$.

Our goal is to design a procurement auction mechanism that satisfies BIC, PAC Compatibility, and IR properties; and minimizes the expected cost of procurement for the learner. We call such an auction mechanism as *Optimal Auction for Data Labeling.* Our design is inspired from the Nobel Prize winning work of Roger Myerson on optimal auction design [Myerson, 1981]. For a comprehensive treatment of this topic, readers are referred to [Krishna, 2002] and [Mishra, 2008].

### 3.2 Characterization of Incentive Compatibility

To design an optimal auction mechanism for data labeling problem, we need to first characterize the space of auction rules that satisfy BIC property. For this, we begin with the following definitions. An allocation rule $a$ is said to be:

- **weakly monotone** (WM) if for every annotator $i$ and for every $\hat{\eta}_{-i} \in I_{-i}$, we have $a_i(\eta_i, \hat{\eta}_{-i}) \geq a_i(\hat{\eta}_i, \hat{\eta}_{-i})$ for all $\eta_i, \hat{\eta}_i \in I_i$, with $\eta_i > \hat{\eta}_i$.
- **weakly monotone in expectation** (WME) if for every annotator $i$ and for every $\eta_i, \hat{\eta}_i \in I_i$ with $\eta_i > \hat{\eta}_i$, we have $\alpha_i(\eta_i) \geq \alpha_i(\hat{\eta}_i)$.

[Myerson, 1981] showed that BIC is characterized by the WME allocation rules in the setting of single object auction. Interestingly, a similar characterization holds in our problem setting also. We state this result as an important theorem.

**Theorem 2** *A Mechanism $\mathcal{M} = (a, p)$ is a BIC mechanism iff (i) the allocation rule $a(\cdot)$ is WME, and (ii) the expected payment rule $\pi(\cdot)$ satisfies:*

$$\pi_i(\eta_i) = \gamma_i + \alpha_i(\eta_i)c(\eta_i) - z_i(\eta_i) \qquad (11)$$

*where $\gamma_i = \pi_i(0) - \alpha_i(0)c(0)$ and $z_i(\eta_i) = \int_0^{\eta_i} \alpha_i(t_i)c'(t_i)dt_i$.*

**Proof:** Suppose $\mathcal{M} = (a, p)$ is a BIC mechanism. Then, we show that the allocation rule is WME and (11) holds. Consider an annotator $i$ and $\eta_i, \hat{\eta}_i \in I_i$ with $\eta_i > \hat{\eta}_i$. Then, it follows from (10) and BIC that $\pi_i(\eta_i) - \alpha_i(\eta_i)c(\eta_i) \geq \pi_i(\hat{\eta}_i) - \alpha_i(\hat{\eta}_i)c(\eta_i)$ and $\pi_i(\hat{\eta}_i) - \alpha_i(\hat{\eta}_i)c(\hat{\eta}_i) \geq \pi_i(\eta_i) - \alpha_i(\eta_i)c(\hat{\eta}_i)$. Adding these two inequalities, we get $[\alpha_i(\eta_i) - \alpha_i(\hat{\eta}_i)][c(\hat{\eta}_i) - c(\eta_i)] \geq 0$. Since $\eta_i > \hat{\eta}_i$ and $c(\cdot)$ is a strictly decreasing function, this implies that $\alpha_i(\eta_i) \geq \alpha_i(\hat{\eta}_i)$ (i.e. $\alpha_i$ is non-decreasing). Now, since $\mathcal{M}$ is BIC, for every $\eta_i, \hat{\eta}_i \in I_i$, we have $U_i(\eta_i) \geq U_i(\hat{\eta}_i) - \alpha_i(\hat{\eta}_i)[c_i(\eta_i) - c_i(\hat{\eta}_i)]$. This is obtained from adding and subtracting $\alpha_i c(\hat{\eta}_i)$ to $U_i(\hat{\eta}_i; \eta_i)$ (Equation (10)) and rearranging the terms. Similarly, switching the roles of $\eta_i$ and $\hat{\eta}_i$, we get $U_i(\hat{\eta}_i) \geq U_i(\eta_i) - \alpha_i(\eta_i)[c_i(\hat{\eta}_i) - c_i(\eta_i)]$. On combining these two inequalities, we get $-\alpha_i(\eta_i)[c_i(\hat{\eta}_i) - c_i(\eta_i)] \leq U_i(\hat{\eta}_i) - U_i(\eta_i) \leq -\alpha_i(\hat{\eta}_i)[c_i(\hat{\eta}_i) - c_i(\eta_i)]$. Now, by dividing with $(\hat{\eta}_i - \eta_i)$ and letting $\hat{\eta}_i \to \eta_i$, the two-sided inequality implies that $-\alpha_i(\cdot)c'(\cdot)$ is the derivative of $U_i(\cdot)$. Since $c'(\cdot)$ is continuous and $\alpha_i(\cdot)$ is non-decreasing, the function $\alpha_i(\cdot)c'(\cdot)$ has finitely many points of discontinuity and hence is Riemann integrable in the interval $[0, 1/3]$. Thus, we have $\int_0^{\eta_i} -\alpha_i(t_i)c'(t_i)dt_i = U_i(\eta_i) - U_i(0)$. On substituting $U_i(\eta_i) = \pi_i(\eta_i) - \alpha_i(\eta_i)c(\eta_i)$ and $U_i(0) = \pi_i(0) - \alpha_i(0)c(0)$, we get (11).

Now, suppose that $a$ is WME and (11) holds. We will show that $\mathcal{M} = (a, p)$ is a BIC mechanism. For any annotator $i$ and any $\eta_i, \hat{\eta}_i \in I_i$, we have $\pi_i(\eta_i) - \pi_i(\hat{\eta}_i) = \alpha_i(\eta_i)c(\eta_i) - \alpha_i(\hat{\eta}_i)c(\eta_i) +$

$\alpha_i(\hat{\eta}_i)\left[c(\eta_i) - c(\hat{\eta}_i)\right] - z_i(\eta_i) + z_i(\hat{\eta}_i)$. Using the facts $\alpha_i(\cdot)$ is non-decreasing and $c'(\cdot) \leq 0$, it can be shown that $\alpha_i(\hat{\eta}_i)\left[c(\eta_i) - c(\hat{\eta}_i)\right] - z_i(\eta_i) + z_i(\hat{\eta}_i) \geq 0$. Therefore, $\pi_i(\eta_i) - \pi_i(\hat{\eta}_i) \geq \alpha_i(\eta_i)c(\eta_i) - \alpha_i(\hat{\eta}_i)c(\eta_i)$ which is the condition for BIC. *Q.E.D*

A similar characterization result can be derived for the DSIC case also. Due to lack of space, we skip the results. Note that Theorem (2) suggests that the learner can only increase the contract size with higher noise rate. This is a bit counter intuitive as the learner is buying more examples from a more noisy annotator (in a relative sense). However, this is essentially the key to enforce truthful elicitation of the noise rates. Even if an annotator misreports higher noise rate to get a bigger size contract, the payment rule would make sure that the additional payment is not enough to cover the cost of labeling the required additional examples. A similar argument holds for the other direction as well.

### 3.3 Optimal Auction Mechanism

We pose the optimization problem of designing the auction mechanism as follows:

$$\min_{a(\cdot), p(\cdot)} \Pi(a, p) = \sum_{i=1}^{n} \int_{0}^{1/3} \pi_i(t_i)\phi_i(t_i)dt_i \quad s.t. \quad (12)$$

$$\alpha_i(\cdot) \text{ is non-decreasing} \quad (13)$$

$$\pi_i(\eta_i) = \gamma_i + \alpha_i(\eta_i)c(\eta_i) - z_i(\eta_i) \; \forall \eta_i \in I_i, \forall i \quad (14)$$

$$\pi_i(\eta_i) \geq \alpha_i(\eta_i)c(\eta_i) \; \forall \eta_i \in I_i, \forall i \quad (15)$$

$$\log(N/\delta) \leq \sum_i a_i(\eta_i, \eta_{-i})\psi(\eta_i) \; \forall(\eta_i, \eta_{-i}) \in I \quad (16)$$

Note that the objective function (12) constitutes the total expected payment made to all the annotators. The constraints (13) and (14) are BIC constraints, (15) is the IR constraint, and (16) is the PAC compatibility constraint. Recall, $c(\cdot)$ is a strictly decreasing function. If (14) is satisfied then (15) will be satisfied iff $\gamma_i \geq 0$ (i.e. $\pi_i(0) \geq \alpha_i(0)c(0)$) $\forall i$. Because our goal is to minimize (12), we must set $\gamma_i = 0$. Then, by setting $\gamma_i = 0$ and using the definition of $\alpha_i(\cdot)$, we can rewrite the objective function after some algebraic manipulations as:

$$\Pi(a, p) = \int_{I} \left(\sum_{i=1}^{n} v_i(x_i)a_i(x)\right) \phi(x)dx \quad (17)$$

where $v_i(\eta_i) = c(\eta_i) - \frac{1-\Phi_i(\eta_i)}{\phi_i(\eta_i)}c'(\eta_i)$ is called as **virtual cost function**. Note that since $c'(\eta_i)$ is negative, $\phi_i(\cdot) > 0$ for all $i$ and for all $\eta_i \in I_i$, the virtual cost function is always non-negative, well defined and is higher than $c(\eta_i)$. Note that (17) is essentially a function of the allocation rule $a(\cdot)$ since $p(\cdot)$ is dictated by $a(\cdot)$ via (14). We need to minimize (17) subject to the constraints (13) and (16). It seems difficult to solve this problem, particularly with the constraint (13) without imposing additional regularity condition. To arrive at this condition, we consider solving (17) by ignoring the constraint (13) momentarily. So, we consider minimizing (17) subject to the constraint (16) alone for the moment. Note, for minimizing (17), it suffices to minimize $\sum_{i=1}^{n} v_i(\eta_i)a_i(\eta)$ for every possible profile $\eta$ subject to the constraint (16). For a fixed $\eta$, this is an integer linear programming (ILP) problem whose approximate solution can be obtained by relaxing the integer constraint and rounding off the optimal solution to the nearest integer value. This approximate solution is a near-optimal way of purchasing examples from noisy annotators for PAC learning (ignoring (13)). By looking at the dual of such a relaxed LP, one can verify that in this near-optimal scheme, the learner should purchase $\lceil \log(N/\delta)/\psi(\eta_{i^*}) \rceil$ number of examples from only that annotator, say $i^*$, for whom the ratio $v_i(\eta_i)/\psi(\eta_i)$ is the minimum. Let us call this rule as the **minimum allocation rule** whose approximation guarantee is given below.

**Theorem 3** *Let ALG be the total cost of purchase incurred by the min allocation rule. Let OPT be the optimal value of the ILP and $m_0$ be non-noisy sample complexity. Then, we must have*

$$ALG \leq OPT + v_{i^*}(\eta_{i^*}) \leq OPT(1 + 1/m_0) \quad (18)$$

**Proof:** The optimal solution of the linear relaxation is always a lower bound on the optimal solution of the ILP. Therefore, we must have $\log(N/\delta)v_{i^*}(\eta_{i^*})/\psi(\eta_{i^*}) \leq OPT$. This implies

$$\begin{aligned} ALG &= v_{i^*}(\eta_{i^*})\lceil \log(N/\delta)/\psi(\eta_{i^*}) \rceil \\ &\leq \log(N/\delta)v_{i^*}(\eta_{i^*})/\psi(\eta_{i^*}) + v_i(\eta_{i^*}) \\ &\leq OPT + v_i(\eta_{i^*}) \quad (19) \end{aligned}$$

To get the other bound, note that the number of examples suggested by the minimum allocation rule is at least as much as $m_0$. Therefore, we must have $\log(N/\delta)/\psi(\eta_{i^*}) \geq m_0$. Thereby, we get:

$$OPT \geq \log(N/\delta)v_{i^*}(\eta_{i^*})/\psi(\eta_{i^*}) \geq m_0 \, v_i(\eta_{i^*}) \quad (20)$$

Substituting the bound on $v_i(\eta_{i^*})$ from (20) into (19) will give us the second term.  Q.E.D

**Regularity Condition:** So far, we considered the approximate optimal auction mechanism design without the constraint (13). Therefore, the minimum allocation rule need not satisfy the WME property. However, it is WME under the regularity condition that $v_i(\cdot)/\psi(\cdot)$ *is a non-increasing function*. Under this condition, as $\eta_i$ increases, the annotator $i$ remains the winner if he/she is already the winner (with an increased contract size) or becomes the winner as per the minimum allocation rule. Therefore, the allocation rule satisfies the WM property (hence, WME). This implies that the allocation rule would give an approximate optimal mechanism satisfying BIC + IR+ PAC compatibility properties. For every $(\eta_i, \eta_{-i})$ and every $i$, the associated payment rule can be given by

$$p_i(\eta_i, \eta_{-i}) = a_i(\eta_i, \eta_{-i})c(\eta_i) - w_i(\eta_i) \qquad (21)$$

where $w_i(\eta_i) = \int_0^{\eta_i} a_i(t_i, \eta_{-i})c'(t_i)dt_i$. One can verify that the corresponding expected payment rule $\pi_i(\cdot)$ satisfies BIC and IR constraints. In fact, it turns out that the minimum allocation rule and the payment rule (21) together satisfy the DSIC property. We skip this proof as it follows the same line of arguments given in the proof of Theorem 2.

**Simplified Payment Rule:** We define for every annotator $i$, the smallest bid value sufficient to win the contract as per the minimum allocation rule as:

$$q_i(\eta_{-i}) = \inf\left\{\hat{\eta}_i \mid \frac{v_i(\eta_i)}{\psi(\eta_i)} \le \frac{v_j(\eta_j)}{\psi(\eta_j)} \;\forall j \ne i\right\} \quad (22)$$

Then, the minimum allocation rule and simplified payment rule can be written as:

$$a_i(\eta) = \begin{cases} \lceil \log(N/\delta)/\psi(\eta_i) \rceil & : \text{if } \eta_i \ge q_i(\eta_{-i}) \\ 0 & : \text{otherwise} \end{cases} \quad (23)$$

$$p_i(\eta) = \begin{cases} \left\lceil \frac{\log(N/\delta)}{\psi(\eta_i)} \right\rceil c(q_i(\eta_{-i})) & : \text{for winner} \\ 0 & : \text{otherwise} \end{cases} \quad (24)$$

Thus, we now have the following mechanism.

**Algorithm 2 (Approx_Mechanism)** *The learner should choose an annotator $i^*$ for whom the score $v_i(\eta_i)/\psi(\eta_i)$ is minimum (breaking ties arbitrarily), and award a contract of supplying $\lceil \log(N/\delta)/\psi(\eta_{i^*}) \rceil$ labeled examples. The learner must pay an amount equal to $c(q_{i^*}(\eta_{-i^*}))$ per example to this annotator, where $q_{i^*}(\eta_{-i^*})$ denotes the smallest noise rate bidding which the winning annotator $i^*$ still stays as the winner. The other annotators are not paid any amount.*

For the winning annotator $i^*$, we have $q_{i^*}(\eta_{-i^*}) \le \eta_{i^*}$ (under the regularity condition). This implies that $c(q_{i^*}(\eta_{-i^*})) \ge c(\eta_{i^*})$. Note that the right hand side of this inequality is the cost involved when $\eta$ is known. Therefore, the learner needs to pay some extra cost to annotators for eliciting the true noise rates. The following theorem is now apparent from the analysis done so far.

**Theorem 4** *Suppose the regularity condition holds. Then,* **Approx_Mechanism** *is an approximate optimal mechanism satisfying DSIC, IR, and PAC compatibility properties. The approximation guarantee of this mechanism is given by $ALG \le OPT + v_{i^*}(\eta_{i^*}) \le OPT(1 + 1/m_0)$.*

Note, DSIC is a preferred condition than BIC and DSIC implies BIC. The above result says that under the regularity condition, a DSIC mechanism comes very close to the optimal BIC mechanism.

## 4 Conclusion

To the best of our knowledge, this is the first paper to model and analyze the problem of acquiring labeled examples from multiple noisy **strategic** annotators for PAC learning. For such a setting, we have proposed an approximate cost optimal auction mechanism for the unknown noise rates scenario, by extending Myerson's optimal auction design framework in a non-trivial manner. As future enhancements, (1) the assumption of finite concept class can be relaxed by making use of VC-dimension, (2) PAC bound can be derived for an improved Weighted MDA (WMDA) algorithm, where we give more importance to the samples from less noisy annotator while computing the loss $L_e(\cdot)$, and (3) one can design better approximate algorithms to solve the underlying ILP problems.

# References

J. Howe. *Crowdsourcing: why the power of the crowd is driving the future of business.* Crown Business, 2008.

L.G. Valiant. A theory of learnable. *Communications of the ACM*, 27:1134–1142, 1984.

D. Angluin and P. Laird. Learning from noisy examples. *Machine Learning*, 2(4):343–370, 1988.

J.A. Aslam and S.E. Decatur. On the sample complexity of noise-tolerant learning. *Information Processing Letters*, 57(4):189–195, 1996.

A. Blum, M. Furst, J. Jackson, M. J. Kearns, Y. Mansour, and S. Rudich. Weakly learning DNF and characterizing statistical query learning using fourier analysis. In *STOC*, 1994.

S. E. Decatur and R. Gennaro. On learning from noisy and incomplete examples. In *COLT*, pages 353–360, 1995.

S. E. Decatur. PAC learning with constant-partition classification noise and applications to decision tree induction. In *Proceedings of the Sixth International Workshop on Artificial Intelligence and Statistics*, pages 147 – 156, 1997.

M. Kearns. Efficient noise-tolerant learning from statistical queries. In *25th ACM Symposium on the Theory of Computing (STOC)*, pages 392 – 401, 1993.

N. Littlestone. Redundant noisy attributes, attribute errors, and linear-threshold learning using winnow. In *Conference on Learning Theory (COLT)*, 1991.

A. P. Dawid and A.M. Skene. Maximum likelihood estimation of observer error-rates using the EM algorithm. *Applied Statistics*, 28(1):20–28, 1979.

V. C. Raykar, S. Yu, L. H. Zhao, A. Jerebko, C. Florin, G. H. Valadez, L. Bogoni, and L. Moy. Supervised learning from multiple experts: Whom to trust when everyone lies a bit. In *International Conference on Machine Learning (ICML)*, 2009.

Y. Yan, G. Hermosillo, R. Rosales, L. Bogoni, G. Fung, L. Moy, M. Schmidt, and J. G. Dy. Modeling annotator expertise: Learning when everybody knows a bit of something. In *International Conference on Artificial Intelligence and Statistics (AISTATS)*, 2010.

P. Donmez, J. Carbonell, and J. Schneider. A probabilistic framework to learn from multiple annotators with time-varying accuracy. In *Proceedings of the SIAM International Conference on Data Mining*, 2010.

L. G. Valiant. Learning disjunctions of conjunctions. In *International Joint Conference on Artificial Intelligence (IJCAI)*, pages 560 – 566, 1985.

M. Kearns and M. Li. Learning in the presence of malicious error. *SIAM Journal on Computing*, 22(4):807–837, 1993.

S.A. Goldman and R.H. Sloan. Can PAC learning algorithms tolerate random attribute noise? *Algorithmica*, 14(1):70–84, 1995.

N. H. Bshouty, N. Eiron, and E. Kushilevitz. PAC learning with nasty noise. *Theoretical Computer Science*, 288(2):255–275, 2002.

O. Dekel and O. Shamir. Vox populi: Collecting high-quality labels from a crowd. In *Conference on Learning Theory (COLT)*, 2009a.

O. Dekel and O. Shamir. Good learners for evil teachers. In *International Conference in Machine Learning (ICML)*, 2009b.

O. Dekel, F. Fischer, and A. D. Procaccia. Incentive Compatible Regression Learning. In *Proceedings of the 19th Annual ACM-SIAM Symposium on Discrete Algorithms (SODA)*, pages 277–286, 2008.

R. Meir, A. D. Procaccia, and J. S. Rosenschein. Strategyproof Classification under Constant Hypotheses: A tale of two functions. In *Proceedings of the 23rd Conference on Artificial Intelligence (AAAI).*, 2008.

R. Meir, A. D. Procaccia, and J. S. Rosenschein. Strategyproof Classification with Shared Inputs. In *Proceedings of the 21st International Joint Conference on Artificial Intelligence (IJCAI)*, 2009.

N. Dalvi, P. Domingos, Mausam, S. Sanghai, and D. Verma. Adversarial Classification. In *International Conference on Knowledge Discovery*

*and Data Mining (KDD)*. ACM, August 22-25 2004.

G. L'Huillier, R. Weber, and N. Figueroa. Online Phishing Classification Using Adversarial Data Mining and Signaling Games. In *International Conference on Knowledge Discovery and Data Mining (KDD)*, 2009.

M. Kantarcioglu, B. Xi, and C. Clifton. A Game Theoretical Framework for Adversarial Learning. In *CERIAS 9th Annual Information Security Symposium*, 2008.

R. B. Myerson. Optimal auction design. *Math. Operations Res.*, 6(1):58–73, Feb. 1981.

P. D. Laird. *Learning from good and bad data*. Kluwer Academic Publishers, Norwell, MA, USA, 1988.

R. Motwani and P. Raghavan. *Randomized Algorithms*. Cambridge University Press, NY, USA, 1995.

V. Krishna. *Auction Theory*. Academic Press, 2002.

D. Mishra. An introduction to mechanism design theory. Unpublished Manuscript. URL: `http://www.isid.ac.in/~dmishra/doc/survey.pdf`, May 2008.